\documentclass{article}
\usepackage{ijcai24}

\usepackage{times}
\usepackage{soul}
\usepackage{url}
\usepackage[hidelinks]{hyperref}
\usepackage[utf8]{inputenc}
\usepackage[small]{caption}
\usepackage{graphicx}
\usepackage{amsmath}
\usepackage{amsthm}
\usepackage{amsfonts}
\usepackage{mathrsfs}
\usepackage{booktabs}
\usepackage{algorithm}
\usepackage{algorithmic}
\usepackage[switch]{lineno}
\usepackage{colortbl}
\usepackage{pifont}
\usepackage[table]{xcolor}
\usepackage{marvosym}

\title{Dual Expert Distillation Network for Generalized Zero-Shot Learning}

\author{
    Anonymous Authors
    \affiliations
    Paper ID: 4799
}

\author{
Zhijie Rao$^{1, \dag}$\and
Jingcai Guo$^{1,2,\dag}$\footnote{Corresponding author: \textit{Jingcai Guo}. $\dag$: Equal contribution.}\and
Xiaocheng Lu$^3$\and
Jingming Liang$^{1,4}$\and
Jie Zhang$^1$\and
Haozhao Wang$^5$\and
Kang Wei$^1$\And
Xiaofeng Cao$^6$
\affiliations
$^1$The Hong Kong Polytechnic University
$^2$PolyU Shenzhen Research Institute\\
$^3$Hong Kong University of Science and Technology
$^4$University of Iowa\\
$^5$Huazhong University of Science and Technology
$^6$Jilin University
\emails
\Letter: jc-jingcai.guo@polyu.edu.hk
}

\begin{document}

\maketitle

\begin{abstract}
Zero-shot learning has consistently yielded remarkable progress via modeling nuanced one-to-one visual-attribute correlation. Existing studies resort to refining a uniform mapping function to align and correlate the sample regions and subattributes, ignoring two crucial issues: 1) the inherent asymmetry of attributes; and 2) the unutilized channel information. This paper addresses these issues by introducing a simple yet effective approach, dubbed \textbf{\underline{D}}ual \textbf{\underline{E}}xpert \textbf{\underline{D}}istillation \textbf{\underline{N}}etwork (DEDN), where two experts are dedicated to coarse- and fine-grained visual-attribute modeling, respectively. Concretely, one coarse expert, namely \textbf{cExp}, has a complete perceptual scope to coordinate visual-attribute similarity metrics across dimensions, and moreover, another fine expert, namely \textbf{fExp}, consists of multiple specialized subnetworks, each corresponds to an exclusive set of attributes. Two experts cooperatively distill from each other to reach a mutual agreement during training. Meanwhile, we further equip DEDN with a newly designed backbone network, i.e., \textbf{\underline{D}}ual \textbf{\underline{A}}ttention \textbf{\underline{N}}etwork (DAN), which incorporates both region and channel attention information to fully exploit and leverage visual semantic knowledge. Experiments on various benchmark datasets indicate a new state-of-the-art. Code is available at \href{https://github.com/zjrao/DEDN}{github.com/zjrao/DEDN}
\end{abstract}

\section{Introduction}

\begin{figure}[t]
    \centering
    \begin{minipage}{\linewidth}
    \includegraphics[width=0.85\textwidth]{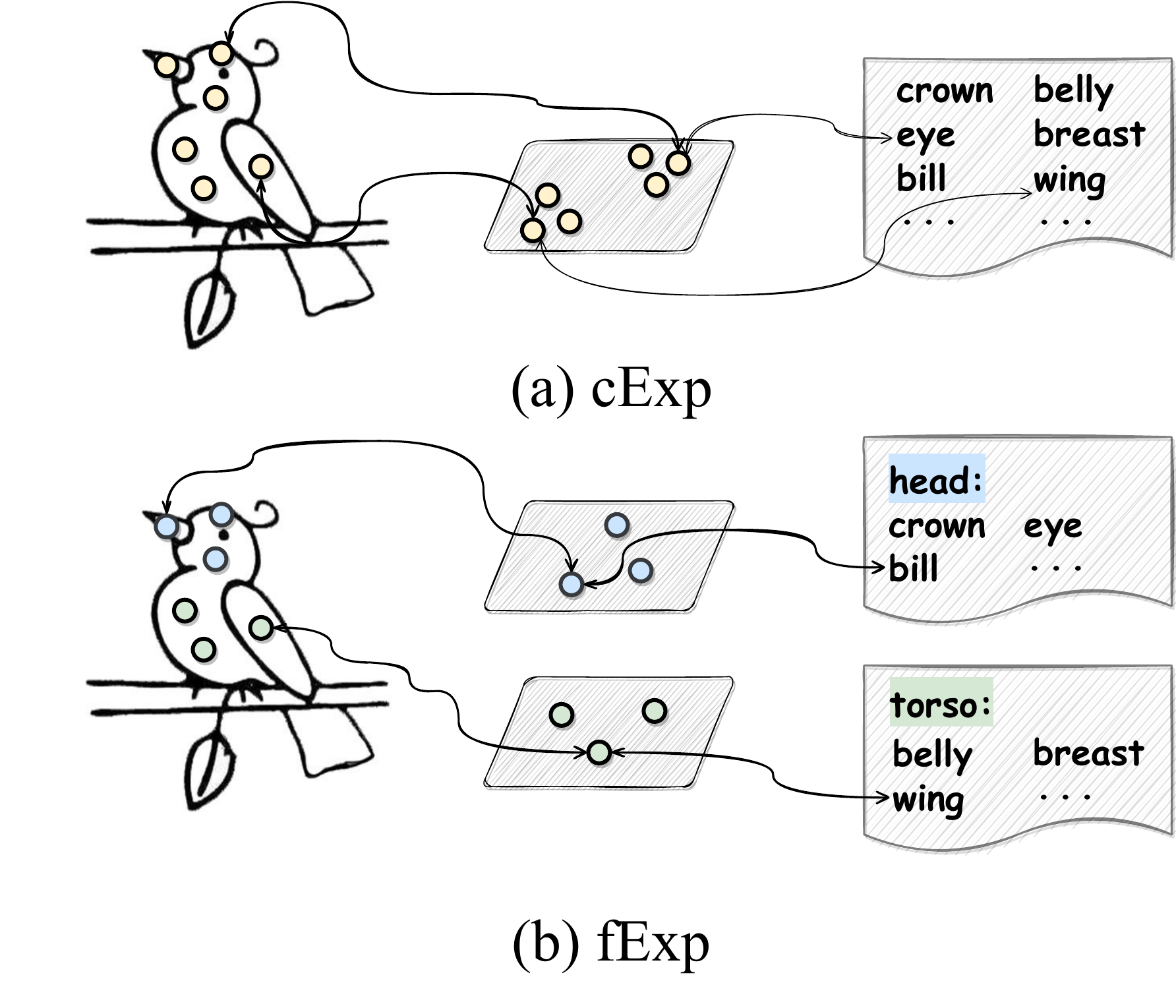}
    \end{minipage}
    \caption{(a) \emph{cExp}, also the common practice in existing works, possesses complete attribute-awareness capability yet lacks the ability to process fine-grained semantic information. (b) \emph{fExp}, which consists of multiple specialized sub-networks, lacks a global perception field.}
    \label{fig:intro}
\end{figure}

Recognizing unknown categories in the open environment is a critical challenge for automatic recognition systems. Zero-Shot Learning (ZSL) \cite{lampert2009learning} that serves as a promising solution has received increasing attention, which is inspired by human text-to-image reasoning capabilities. The objective of ZSL is to transfer the visual knowledge of seen classes to the unseen domain by virtue of shared semantic information, thus empowering the model to recognize the unseen classes. More trickily, Generalized Zero-Shot Learning (GZSL) \cite{chao2016empirical} requires recognizing samples from both seen and unseen classes in the inference phase.

Mainstream studies broadly follow two routes, generative \cite{xian2018feature}\cite{xie2022leveraging}\cite{Li2023boosting} and embedding techniques \cite{zhang2017learning}\cite{liu2020attribute}\cite{chen2021hsva}, where most of the schemes are devoted to mining and constructing class-wise visual-attribute relations. To strengthen the fine-grained perceptual capabilities of the model, recent research has invested considerable effort into modeling local-subattribute correlations \cite{xie2019attentive}\cite{huynh2020fine}\cite{xu2020attribute}. The motivation is to build a refined pairwise relation map via searching and binding subattributes and the corresponding region visual features (Figure \ref{fig:intro} (a)). Despite their contribution to boosting performance, the inherent asymmetry of attributes remains undiscussed, and the channel information is not fully exploited. 

The asymmetry of attributes stems from the fact that 1) the semantic dimensions between attributes are heterogeneous or even antagonistic. Take the SUN dataset \cite{patterson2012sun} as an example, where 38 attributes (\emph{studying, playing, etc.}) describe the function of one scene, while 27 attributes (\emph{trees, flowers, etc.}) describe the entities in the scene. It can be obviously observed that the former are abstract and global, while the latter are concrete and local; 2) the visual features corresponding to attributes are intertwined. For example, neighboring regions tend to be more semantically similar, a phenomenon that is exacerbated by the local information fusion mechanism of the convolutional kernel, which leads to difficulties in accurately locating fine-grained attributes such as \emph{head, crown}, and so on. 

In this paper, we revisit the task of modeling visual-attribute relations from the perspective of attribute annotations. Given the inherent complexity of attribute descriptions, existing learning paradigms are virtually forcing a single model to undertake a multi-objective hybrid task, which is ideally appealing yet empirically challenging. Naturally, we employ the idea of divide-and-conquer to release the pressure of a single model. We meticulously decompose the hybrid task into multiple subtasks, i.e., dividing the attributes into multiple disjoint clusters and assigning specialized learnable networks to them. Our approach is referred to as, Dual Expert Distillation Network, abbreviated DEDN. As shown in Figure \ref{fig:intro}, our approach sets up two experts. \emph{cExp}, in line with common practices, is equipped with complete attribute perception capability to harmonize holistic visual-attribute measure results. \emph{fExp}, consists of multiple subnetworks, where each subnetwork is only responsible for capturing the characteristics of a specific attribute cluster. During the training phase, we encourage the two to learn cooperatively to compensate for their respective deficiencies in a mutually distilling manner. The decision results of the two experts are combined for final inference.

For the issue of underutilized channel information, we design a novel attention network, Dual Attention Network (DAN), as the backbone. DAN employs a dual-attention mechanism that fully exploits the potential semantic knowledge of both regions and channels to facilitate more precise visual-attribute correlation metrics. To further boost performance, we present Margin-Aware Loss (MAL) as the training loss function to address the confidence imbalance between seen and unseen classes.

Our contributions are summarized below:

\begin{itemize}

    \item We rethink the issue of modeling visual-attribute relations from the perspective of attribute annotations and point out that the inherent complexity of attributes is one of the major bottlenecks. We propose a simple yet effective strategy of establishing two experts working on distinct attribute perception scopes to learn and infer collaboratively in a complementary manner.

    \item We present a novel attention network, dubbed DAN, which incorporates both region and channel attention information to better capture correlations between visuals and attributes. Furthermore, a new learning function named MAL is designed to balance the confidence of seen and unseen classes.

    \item We conduct extensive experiments on mainstream evaluation datasets, and the results show that the proposed method effectively improves the performance.
    
\end{itemize}

\section{Related Work}

In ZSL/GZSL, attributes are the only ties that bridge seen and unseen classes, hence exploring and constructing the link between visuals and attributes is a core subject. Existing methods fall into class-wise visual-attribute modeling, which treats both visual features and attribute vectors as a whole, and regional visual-subattribute modeling, which seeks to explore the correlation between local visual information and subattributes.

\subsection{Class-wise Visual-Attribute Modeling}

Mainstream researches broadly follow two technical routes, generative and embedding techniques. Generative techniques utilize the latent distribution fitting ability of generative models such as GAN and VAE to implicitly learn the relationship between attributes and categories to construct hallucinatory samples of unseen classes \cite{xian2018feature}\cite{verma2018generalized}\cite{felix2018multi}\cite{li2019leveraging}\cite{vyas2020leveraging}\cite{keshari2020generalized}\cite{xie2022leveraging}\cite{Li2023boosting}. The technical bottleneck of this route is the poor realism of the hallucinatory samples, thus many studies incorporate other techniques such as meta-learning \cite{yu2020episode}, representation learning \cite{li2021generalized}\cite{chen2021semantics}\cite{chen2021free}\cite{han2021contrastive}\cite{kong2022compactness}, etc. for joint training. Embedding techniques aim at projecting visual and attribute features to a certain space, from which the most similar semantic information is searched. In general, embedding techniques are categorized into three directions: visual-to-attribute space \cite{changpinyo2016synthesized}\cite{kodirov2017semantic}\cite{liu2020attribute}\cite{chen2022transzero}, attribute-to-visual space \cite{zhang2017learning}\cite{annadani2018preserving}, and common space \cite{liu2018generalized}\cite{jiang2019transferable}. Researchers in the first two directions invest considerable effort in designing robust mapping functions to cope with domain shift and out-of-distribution generalization problems. The third direction centers on finding a suitable semantic space. Class-level visual-attribute modeling lacks the fine-grained perceptual ability to respond to interactions between local visual features and subattributes.

\begin{figure*}[t]
    \centering
    \begin{minipage}{\linewidth}
    \includegraphics[width=0.99\textwidth]{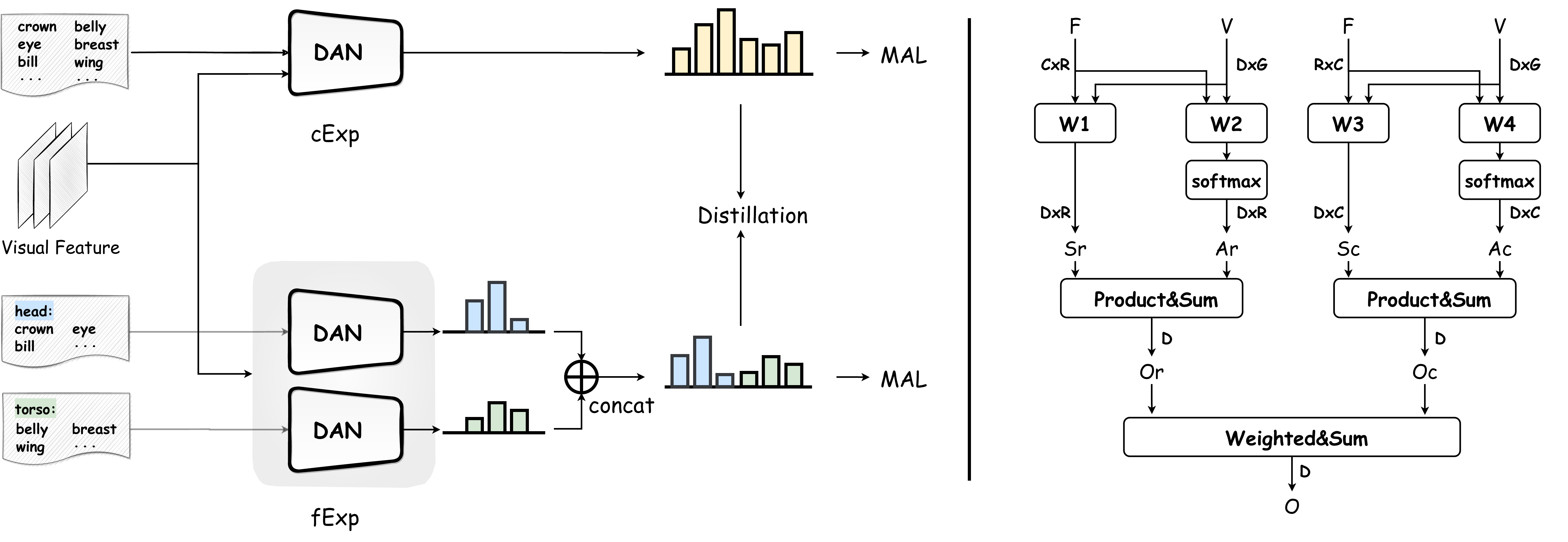}
    \end{minipage}
    \caption{\textbf{Left:} \emph{cExp} possesses the scope of a holistic attribute set, while \emph{fExp} consists of multiple sub-networks, each of which is responsible for the prediction of only partial attributes. We concatenate all outputs of subnetworks as the final result of \emph{fExp}. Then, distillation loss is implemented to facilitate joint learning. \textbf{Right:} The architecture of DAN.}
    \label{fig:method}
\end{figure*}

\subsection{Region-wise Visual-Attribute Modeling}

Region-wise modeling is a promising direction in embedding techniques. Unlike other embedding approaches, region-wise modeling focuses on the correlation between local information and subattributes to build more detailed mapping functions. Models based on attention mechanisms are the dominant means in this direction, motivated by training models to search for corresponding visual features based on semantic vectors. Recent approaches include feature-to-attribute attention networks \cite{xie2019attentive}\cite{huynh2020fine}, bidirectional attention networks \cite{chen2022msdn}, and multi-attention networks \cite{zhu2019semantic}. In addition, some studies resort to prototype learning, where the goal is to explicitly learn the corresponding prototypical visual features of individual subattributes, thus aiding the model's judgment \cite{xu2020attribute}\cite{wang2021dual}. Further, modeling the topological structure between regional features with the help of graph convolution techniques also yields promising results \cite{xie2020region}\cite{guo2023graph}. While the main idea of these approaches is to design appropriate attention networks or regularization functions, ignoring the inherent complexity of attribute annotations, we provide a new perspective to think about the visual-attribute modeling problem. In addition, existing region-attribute methods, although achieving good results, neglect the utilization of channel information, and we design a new attention network that utilizes both region and channel information.

\section{Methodology}

\subsection{Preliminary}

Following previous studies \cite{chen2022msdn}\cite{Li2023boosting}, we adopt a fixed feature extractor, ResNet-101 \cite{he2016deep}, to extract visual features. Suppose $\mathcal{D}^s= \{(F_i^s, Y_i^s)\}$ denotes the seen classes, where $F_i^s$ is the visual feature and $Y_i^s$ denotes its label. Note that $F\in\mathbb{R}^{C\times H\times W}$, where $C, H, W$ are the channel number, height, and width, respectively. Similarly have $\mathcal{D}^u= \{(F_i^u, Y_i^u)\}$ to denote the unseen classes. Normally, the visual features of the unseen classes are not accessible during the training phase. Alternatively, we have the shared attribute $A\in\mathbb{R}^{K\times D}$, where $K$ denotes the total number of categories, and $D$ denotes the number of attributes. Also, we use the semantic vectors of each attribute learned by GloVe, denoted by $V\in\mathbb{R}^{D\times G}$, where $G$ denotes the dimension of the vector.

\subsection{Overview}

Our approach is shown in Figure \ref{fig:method} (Left). First, we disassemble the attribute set into multiple clusters based on their characteristics. Then the attribute vectors and the visual feature are fed into \emph{cExp} and \emph{fExp} simultaneously. \emph{cExp} directly computes the scores of all attributes on that visual feature, while the scores of \emph{fExp} are obtained by combining the computation results of each subnetwork. We constrain the two to learn from each other using distillation loss. Meanwhile, we introduce DAN as the backbone and MAL as the optimization objective.

\subsection{Dual Attention Network}

Firstly we introduce the proposed novel backbone network, Dual Attention Network (DAN). Mining and constructing relations between visual features and attributes is crucial for zero-shot learning. Recently many works have been devoted to modeling the association between regions and attributes, such as attention-based approaches \cite{xie2019attentive}\cite{huynh2020fine}\cite{chen2022msdn} and prototype-based techniques \cite{xu2020attribute}\cite{wang2021dual}. However, these methods only focus on the semantic information of regions and ignore the role of channels. Therefore, DAN incorporates both the attention information of regions and channels to promote the efficacy of the model in utilizing visual features.

As shown in Figure \ref{fig:method} (Right), DAN contains two parallel components that model region-attribute and channel-attribute relations, respectively. We first introduce the region-attribute component. We have visual features $F\in\mathbb{R}^{C\times H\times W}$, which is flattened to $F\in\mathbb{R}^{C\times R}$, where $R=H\times W$ denotes the number of regions. Let $W_1, W_2\in\mathbb{R}^{G\times C}$ denote two learnable matrices. $W_1$ maps the attribute vectors to the visual space and computes their similarity. The formula is expressed as:
\begin{equation}
    S_r = VW_1F,
\end{equation}
where $S_r\in\mathbb{R}^{D\times R}$ represents the score obtained for each attribute on each region. $W_2$ is in charge of computing the attention weights to encourage the model to focus on the region-attribute pairs with the highest similarity. The formula is expressed as:
\begin{equation}
    A_r = \frac{VW_2F}{\sum_{r\in R}VW_2F_r},
\end{equation}
where $A_r\in\mathbb{R}^{D\times R}$ denote the normalized weight obtained by softmax. Then we naturally get the weighted matrix of scores, represented as:
\begin{equation}
    O_r = \sum_R S_r\times A_r,
\end{equation}
where $O_r\in\mathbb{R}^{D}$ represents the similarity score obtained for each attribute on a visual feature. 

Next, we introduce the channel-attribute section, which has a similar principle. We have the scaled visual feature $F\in\mathbb{R}^{R\times C}$ and $W_3, W_4\in\mathbb{R}^{G\times R}$. Then $W_3$ is charged with calculating the similarity score obtained by the attribute on each channel, formulated as:
\begin{equation}
    S_c = VW_3F,
\end{equation}
where $S_c\in\mathbb{R}^{D\times C}$. And $W_4$ computes its attention weights:
\begin{equation}
    A_c = \frac{VW_4F}{\sum_{c\in C}VW_4F_c},
\end{equation}
where $A_c\in\mathbb{R}^{D\times C}$. Finally, we get the weighted score map:
\begin{equation}
    O_c = \sum_C S_c\times A_c,
\end{equation}
where $O_c\in\mathbb{R}^{D}$. We expect the final scores of attributes from different scale features to be consistent, i.e., semantic consistency. Therefore we employ $\mathcal{L}_{align}$, which contains a Jensen-Shannon Divergence (JSD) and a Mean Squared Error, to align the outputs of both, formulated as:
\begin{equation}
    \mathcal{L}_{align}=\frac{1}{2}(\mathcal{L}_{KL}(O_r||O_c)+\mathcal{L}_{KL}(O_c||O_r))+||O_r-O_c||_2^2,
\end{equation}
where $\mathcal{L}_{KL}$ denotes Kullback-Leibler Divergence. In the inference phase, we use the weighted sum of $O_r$ and $O_c$ as the final output, expressed as:
\begin{equation}
    O=\lambda_{rc}\times O_{r}+(1-\lambda_{rc})\times O_{c},
\end{equation}
where $\lambda_{rc}$ is a hyperparameter.

\subsection{Dual Expert Distillation Network}

Despite the fact that DAN enhances the modeling capability of the network, it is extremely challenging for a single model to simultaneously handle attributes with different semantic dimensions as well as visual features with different granularities. To this end, we propose the Dual Expert Distillation Network (DEDN) to alleviate the pressure on a single network (Figure \ref{fig:method} (left)). \emph{cExp} is set up with a complete attribute-aware scope as in conventional practice. Specifically, the input of \emph{cExp} is the semantic vectors of all attributes, and the output is the similarity scores of all attributes. Denote \emph{cExp} by $\phi_{ec}=\{W_1^{ec}, W_2^{ec}, W_3^{ec}, W_4^{ec}\}$, the output is defined as:
\begin{equation}
    O_{ec}=\phi_{ec}(V, F),
\end{equation}
where $O_{ec}\in\mathbb{R}^D$ and $V\in\mathbb{R}^{D\times G}$.

\emph{fExp} consists of multiple subnetworks, each focusing on a specific attribute cluster. At first, we elaborate on how the attribute clusters are divided. Since attribute annotations are manually labeled based on semantics, they are inherently clustered in nature. For example, in the SUN dataset \cite{patterson2012sun}, the top 38 prompts are used to describe the scene function. Therefore, it is easy to perform the division by human operation, Chat-GPT \cite{radford2018improving}, or clustering algorithm. It requires a trivial amount of effort but is worth it. 

Assuming that the attribute set is divided into $Q$ disjoint clusters, i.e. $V=\{V_1\in\mathbb{R}^{D_1\times G}, V_2\in\mathbb{R}^{D_2\times G}, ..., V_Q\in\mathbb{R}^{D_Q\times G}\}$, where $D_1+D_2+...+D_Q=D$. Accordingly, there are $Q$ subnetworks for \emph{fExp} to handle these attribute clusters one-to-one. Let $\phi_{ef}=\{\phi_{ef}^1, \phi_{ef}^2, ..., \phi_{ef}^Q\}$ denotes \emph{fExp}, then the output is defined as:
\begin{equation}
    O_{ef}=\phi_{ef}^1(V_1, F)\oplus\phi_{ef}^2(V_2, F)\oplus...\oplus\phi_{ef}^Q(V_Q, F),
\end{equation}
where $\oplus$ denotes concat operation.

After that, we calculate the score of each category for training and inference. Specifically, we compute the similarity with the output of the expert and the attributes of each category, defined as:
\begin{equation}
    P_{ec}=O_{ec}A^\mathsf{T}, P_{ef}=O_{ef}A^\mathsf{T},
\end{equation}
where $P_{ec}, P_{ef}\in\mathbb{R}^{K}$. To facilitate cooperative learning between two expert networks, we introduce distillation loss to constrain their semantic consistency. Concretely, the distillation loss contains a Jensen-Shannon Divergence (JSD) and a Mean Squared Error, defined as:
\begin{equation}
    \mathcal{L}_{distill}=\frac{1}{2}(\mathcal{L}_{KL}(P_{ec}||P_{ef})+\mathcal{L}_{KL}(P_{ef}||P_{ec}))+||P_{ec}-P_{ef}||_2^2.
\end{equation}

\subsection{Margin-Aware Loss}

Once the category scores are obtained, the network is optimized by using the cross-entropy loss, which is formulated as:
\begin{equation}
    \mathcal{L}_{ce}=-\log\frac{\exp(P_{ec}^y)}{\sum_{y_i}^K\exp(P_{ec}^{y_i})},
\end{equation}
where $y$ is the ground truth. The loss of $P_{ef}$ ditto. Note that we next narrate with $P_{ec}$ only, and the principle is the same for $P_{ef}$.

Due to the lack of access to samples from the unseen classes during the training phase, the scores of the unseen classes are relatively low and thus cannot compete with the seen classes in GZSL. To address this problem, the common practice \cite{huynh2020fine}\cite{chen2022msdn} is to add a margin to the scores:
\begin{equation}
    PM_{ec} = [P_{ec}^1-\epsilon, ..., P_{ec}^N-\epsilon, P_{ec}^{N+1}+\epsilon,..., P_{ec}^K+\epsilon],
\end{equation}
where $\epsilon$ is a constant, $P_{ec}^1\sim P_{ec}^N$ are seen classes score, and $P_{ec}^{N+1}\sim P_{ec}^K$ are unseen classes score. However, this method leads to misclassification of seen classes that would otherwise be correctly predicted. In order to maintain the correctness of the predicted classes while enhancing the competitiveness of the unseen classes. We propose Margin-Aware Loss (MAL), which takes the form:
\begin{equation}
    \resizebox{0.99\linewidth}{!}{$
    \mathcal{L}_{mal}=-\log\frac{\exp(P_{ec}^y-2\epsilon)}{\exp(P_{ec}^y-2\epsilon)+\sum_{y_i\neq y}^{\mathcal{S}}\exp(P_{ec}^{y_i}+\epsilon)+\sum^{\mathcal{U}}\exp(P_{ec}^{y_i})}$
    },
\end{equation}
where $\mathcal{S}, \mathcal{U}$ denote seen and unseen classes, respectively. In contrast to the cross-entropy loss, MAL reactivates the confidence of the predicted class to ensure that it stays ahead in the margin-processed scores, while suppressing the confidence of the other seen classes to ensure the competitiveness of the unseen classes.

\subsection{Summarize}

In the training phase, the basic training loss of \emph{cExp} stems from the classification and the alignment loss, which is expressed as:
\begin{equation}
\mathcal{L}_{ec}=\mathcal{L}_{mal}^{ec}+\beta\mathcal{L}_{align}^{ec},
\end{equation}
where $\beta$ is a hyperparameter. Similarly, we have the basic training loss of \emph{fExp}:
\begin{equation}
\mathcal{L}_{ef}=\mathcal{L}_{mal}^{ef}+\beta\mathcal{L}_{align}^{ef}.
\end{equation}
Then the final loss is obtained from the combination of basic losses and distillation loss, denoted as:
\begin{equation}
    \mathcal{L}_{DEDN}=\mathcal{L}_{ec}+\mathcal{L}_{ef}+\gamma\mathcal{L}_{distill},
\end{equation}
where $\gamma$ is a hyperparameter.

In the inference phase, the recommendations of the two experts are combined and used for final judgment. The predicted result is expressed as:
\begin{equation}
    \arg\max\lambda_{e}\times P_{ec}+(1-\lambda_{e})\times P_{ef},
\end{equation}
where $\lambda_{e}$ is a hyperparameter.

\section{Experiments}

\noindent\textbf{Datasets.} We conduct extensive experiments on three benchmark datasets to verify the effectiveness of the method, including CUB (Caltech UCSD Birds 200) \cite{wah2011caltech}, SUN (SUN Attribute) \cite{patterson2012sun}, and AWA2 (Animals with Attributes 2) \cite{xian2017zero}. We split all datasets following \cite{xian2017zero}. CUB comprises 200 bird species totaling 11,788 image samples, of which 50 categories are planned as unseen classes. We use class attributes for fair comparison, which contain 312 subattributes. SUN has a sample of 717 different scenes totaling 14,340 images, where 72 categories are unseen classes. Attribute annotations are 102-dimensional. AWA2 includes 50 classes of assorted animals totaling 37,322 samples, of which 10 categories are considered unseen classes. Its number of attributes is 85.

\hspace*{\fill}\

\noindent\textbf{Evaluation Protocols.} We perform experiments in both the Zero-Shot learning (ZSL) and Generalized Zero-Shot learning (GZSL) settings. For ZSL, we employ top-1 accuracy to evaluate the performance of the model, denoted as \textbf{T}. For GZSL, we record the accuracy for both seen classes, and unseen classes, denoted as \textbf{S}, and \textbf{U}, respectively. We also record the harmonic mean \textbf{H}, which is computed as, $\mathrm{H}=(2\times\mathrm{S}\times\mathrm{U})/(\mathrm{S}+\mathrm{U})$.

\hspace*{\fill}\

\noindent\textbf{Implementation Details.} For a fair comparison, we use the fixed ResNet-101 \cite{he2016deep} without finetune as the feature extractor. We set the batch size to 50 and the learning rate to 0.0001. The RMSProp optimizer with the momentum set as 0.9 and weight decay set as 1e-4 is employed. For hyperparameters, $[\beta, \gamma]$ are fixed to [0.001, 0.1]. We empirically set $[\lambda_{rc}, \lambda_e]$ to [0.8, 0.9] for CUB, [0.95, 0.3] for SUN, [0.8, 0.5] for AWA2. Subsequent experimental analyses show that the performance of our method has \textbf{low sensitivity} to hyperparameters. For attribute clusters, we classify attribute sets according to their characteristics, and the results are shown in Table \ref{tab:attribute_cluster}.

\begin{table}[t]
    \centering
    \begin{tabular}{llllll}
         \toprule
         \multicolumn{2}{c}{\textbf{CUB}} &  \multicolumn{2}{c}{\textbf{SUN}} & \multicolumn{2}{c}{\textbf{AWA2}}\\
         \cmidrule(r){1-2}\cmidrule(r){3-4}\cmidrule(r){5-6}
         \#Des. & \#Num. & \#Des. & \#Num. & \#Des. & \#Num. \\
         \hline
         head & 112 & function & 38 & texture & 18\\
         torso & 87 & instance & 27 & organ & 14\\
         wing & 24 & environ. & 17 & environ. & 13\\
         tail & 40 & light & 20 & abstract & 40\\
         leg & 15 & & & & \\
         whole & 34 & & & & \\
         \bottomrule
    \end{tabular}
    \centering
    \caption{Manual division of attribute clusters. \emph{Des.} (description) indicates the criteria for classification. \emph{Num.} (number) is the size of the attribute cluster. \emph{environ}: environment.}
    \label{tab:attribute_cluster}
\end{table}

\subsection{Compared with State-of-the-arts}

To evaluate the performance of the proposed method, we compare it with the state-of-the-art various methods. Generative methods: f-CLSWGAN (CVPR $'18$) \cite{xian2018feature}, f-VAEGAN-D2 (CVPR $'19$) \cite{xian2019f}, TF-VAEGAN (ECCV $'20$) \cite{narayan2020latent}, E-PGN (CVPR $'20$) \cite{yu2020episode}, CADA-VAE (CVPR $'19$) \cite{schonfeld2019generalized}, FREE (ICCV $'21$) \cite{chen2021free}, SDGZSL (ICCV $'21$) \cite{chen2021semantics}, CE-GZSL (CVPR $'21$) \cite{han2021contrastive}, VS-Boost (IJCAI $'23$) \cite{Li2023boosting}; Embedding methos: LFGAA (ICCV $'19$) \cite{liu2019attribute}, APN (NeurIPS $'20$) \cite{xu2020attribute}, DCN (NeurIPS $'18$) \cite{liu2018generalized}, HSVA (NeurIPS $'21$) \cite{chen2021hsva}; Region-Attribute modeling: SGMA (NeurIPS $'19$) \cite{zhu2019semantic}, AREN (CVPR $'19$) \cite{xie2019attentive}, DAZLE (CVPR $'20$) \cite{huynh2020fine}, MSDN (CVPR $'22$) \cite{chen2022msdn}.

\begin{table*}[t]
    \centering
    \begin{tabular}{l c c c c c c c c c c c c c}
        \toprule
         & & \multicolumn{4}{c}{\textbf{CUB}} & \multicolumn{4}{c}{\textbf{SUN}} & \multicolumn{4}{c}{\textbf{AWA2}} \\
        \cmidrule(r){3-6} \cmidrule(r){7-10} \cmidrule(r){11-14}
        METHOD & ROUTE & T & U & S & H & T & U & S & H & T & U & S & H\\
        \hline
        f-CLSWGAN & Gen. & 57.3 & 43.7 & 57.7 & 49.7 & 60.8 & 42.6 & 36.6 & 39.4 & 68.2 & 57.9 & 61.4 & 59.6\\
        f-VAEGAN-D2 & Gen. & 61.0 & 48.4 & 60.1 & 53.6 & 64.7 & 45.1 & 38.0 & 41.3 & 71.1 & 57.6 & 70.6 & 63.5\\
        TF-VAEGAN & Gen. & 64.9 & 52.8 & 64.7 & 58.1 & \underline{66.0} & 45.6 & \textcolor{blue!60}{\textbf{40.7}} & 43.0 & 72.2 & 59.8 & 75.1 & 66.6\\
        E-PGN & Gen. & 72.4 & 52.0 & 61.1 & 56.2 & - & - & - & - & \underline{73.4} & 52.6 & 83.5 & 64.6\\
        CADA-VAE & Gen. & 59.8 & 51.6 & 53.5 & 52.4 & 61.7 & 47.2 & 35.7 & 40.6 & 63.0 & 55.8 & 75.0 & 63.9\\
        FREE & Gen. & - & 55.7 & 59.9 & 57.7 & - & 47.4 & 37.2 & 41.7 & - & 60.4 & 75.4 & 67.1\\
        SDGZSL & Gen. & 75.5 & 59.9 & 66.4 & 63.0 & 62.4 & 48.2 & 36.1 & 41.3 & 72.1 & 64.6 & 73.6 & 68.8\\
        CE-GZSL & Gen. & \underline{77.5} & 63.9 & 66.8 & 65.3 & 63.3 & 48.8 & 38.6 & 43.1 & 70.4 & 63.1 & 78.6 & 70.0\\
        VS-Boost & Gen. & \textcolor{blue!60}{\textbf{79.8}} & 68.0 & 68.7 & \underline{68.4} & 62.4 & 49.2 & 37.4 & 42.5 & - & \underline{67.9} & 81.6 & \textcolor{blue!60}{\textbf{74.1}}\\
        SGMA & Emb.$\dagger$ & 71.0 & 36.7 & 71.3 & 48.5 & - & - & - & - & 68.8 & 37.6 & 87.1 & 52.5\\
        AREN & Emb.$\dagger$ & 71.8 & 38.9 & \underline{78.7} & 52.1 & 60.6 & 19.0 & 38.8 & 25.5 & 67.9 & 15.6 & \underline{92.9} & 26.7\\
        LFGAA & Emb. & 67.6 & 36.2 & \textcolor{blue!60}{\textbf{80.9}} & 50.0 & 61.5 & 18.5 & \underline{40.0} & 25.3 & 68.1 & 27.0 & \textcolor{blue!60}{\textbf{93.4}} & 41.9\\
        DAZLE & Emb.$\dagger$ & 66.0 & 56.7 & 59.6 & 58.1 & 59.4 & \underline{52.3} & 24.3 & 33.2 & 67.9 & 60.3 & 75.7 & 67.1\\
        APN & Emb. & 72.0 & 65.3 & 69.3 & 67.2 & 61.6 & 41.9 & 34.0 & 37.6 & 68.4 & 57.1 & 72.4 & 63.9\\
        DCN & Emb. & 56.2 & 28.4 & 60.7 & 38.7 & 61.8 & 25.5 & 37.0 & 30.2 & 65.2 & 25.5 & 84.2 & 39.1\\
        HSVA & Emb. & 62.8 & 52.7 & 58.3 & 55.3 & 63.8 & 48.6 & 39.0 & \underline{43.3} & - & 59.3 & 76.6 & 66.8\\
        MSDN & Emb.$\dagger$ & 76.1 & \underline{68.7} & 67.5 & 68.1 & 65.8 & 52.2 & 34.2 & 41.3 & 70.1 & 62.0 & 74.5 & 67.7\\
        \rowcolor{blue!5}
        DEDN(\textbf{Ours}) & Emb. & 77.4 & \textcolor{blue!60}{\textbf{70.9}} & 70.0 & \textcolor{blue!60}{\textbf{70.4}} & \textcolor{blue!60}{\textbf{67.4}} & \textcolor{blue!60}{\textbf{54.7}} & 36.0 & \textcolor{blue!60}{\textbf{43.5}} & \textcolor{blue!60}{\textbf{75.8}} & \textcolor{blue!60}{\textbf{68.0}} & 76.5 & \underline{72.0}\\ 
        \bottomrule
    \end{tabular}
    \centering
    \caption{Comparison with state-of-the-art methods (\%). \emph{Gen.} denotes generative method and \emph{Emb.} denotes embedding method. $\dagger$ denotes the region-attribute modeling method. The best and second-best results are highlighted in \textcolor{blue!60}{\textbf{blue}} and \underline{underlined}, respectively.}
    \label{tab:mainresults}
\end{table*}

\begin{table*}[htbp]
    \centering
    \begin{tabular}{l c c c c c c c c c c c c}
        \toprule
         & \multicolumn{4}{c}{\textbf{CUB}} & \multicolumn{4}{c}{\textbf{SUN}} & \multicolumn{4}{c}{\textbf{AWA2}} \\
        \cmidrule(r){2-5} \cmidrule(r){6-9} \cmidrule(r){10-13}
        SETTING & T & U & S & H & T & U & S & H & T & U & S & H\\
        \hline
        cExp w/o $\mathcal{L}_{distill}$ & 74.6 & 62.4 & 71.4 & 66.6 & 64.0 & 41.6 & 35.7 & 38.4 & 71.1 & 62.8 & 78.8 & 69.9\\
        fExp w/o $\mathcal{L}_{distill}$ & 75.5 & 68.1 & 67.9 & 68.0 & 64.0 & 42.8 & 35.5 & 38.7 & 71.1 & 62.9 & 79.1 & 70.1\\
        DEDN w/o $\mathcal{L}_{distill}$ & 75.7 & 66.7 & 70.7 & 68.6 & 65.2 & 47.3 & 35.0 & 40.3 & 72.1 & 63.8 & 79.3 & 70.7\\
        DAN w/o CA$*$ & 77.0 & 58.7 & 73.6 & 65.3 & 65.8 & 48.5 & 34.6 & 40.4 & 74.6 & 61.7 & \textbf{79.8} & 69.6\\
        DEDN w/o $\mathcal{L}_{mal}$ & 75.8 & \textbf{73.2} & 62.5 & 67.4 & 66.0 & \textbf{56.5} & 34.3 & 42.7 & 73.1 & 66.5 & 72.4 & 69.3\\
        DAN w/o $\mathcal{L}_{align}$ & \textbf{77.6} & 63.3 & \textbf{72.8} & 67.7 & 65.5 & 47.5 & 35.3 & 40.5 & 74.6 & 64.8 & 76.8 & 70.3\\
        DEDN(full) & 77.4 & 70.9 & 70.0 & \textbf{70.4} & \textbf{67.4} & 54.7 & \textbf{36.0} & \textbf{43.5} & \textbf{75.8} & \textbf{68.0} & 76.5 & \textbf{72.0} \\ 
        \bottomrule
    \end{tabular}
    \centering
    \caption{Ablation Study (\%). \emph{w/o} denotes remove the module. \emph{CA}$*$ denotes channel attention. The best result is highlighted in {\textbf{bold}}.}
    \label{tab:ablation}
\end{table*}

The experimental results are shown in Table 1. Our method achieves the best performance in seven metrics and second place in one metric. For Generalized Zero-Shot Learning (GZSL), we beat VS-Boost by 2\% in the H-score of CUB, a fine-grained bird dataset whose attribute annotations possess explicit correspondences to visual features. It demonstrates the superiority of the proposed method for fine-grained modeling. On the SUN and AWA2 datasets, we obtain the best and second-best results in H-score, respectively. These two datasets have fewer attributes and contain complex semantic dimensions, including abstract, concrete, etc. The experimental results demonstrate the effectiveness of the proposed method in deconstructing complex tasks to alleviate the modeling pressure of a single network. In addition, the U-scores of our method on all three datasets are well ahead of the others, demonstrating that the proposed method effectively captures the relationship between attributes and visuals to generalize to unseen classes.

For Zero-Shot Learning (ZSL), we achieve the highest top-1 accuracy on the SUN and AWA2 datasets, as well as competitive performance on CUB. Specifically, our method outperforms TF-VAEGAN by 1.4\% on the SUN dataset. On AWA2, we have a 2.4\% lead relative to the second-place E-PGN. The experimental results validate the superiority of the proposed method. Notably, our method achieves far better results than existing region-attribute modeling methods in both ZSL and GZSL settings, which implies the potential of attribute intrinsic asymmetry and channel information is not fully exploited.

\subsection{Ablation Study}

To evaluate the role of each module, we perform a series of ablation experiments. The results of the experiments are shown in Table \ref{tab:ablation}. Comprehensively, removing any of the modules leads to different degrees of performance degradation, verifying the rationality and necessity of the design of each module. Concretely, it is observed that the performance of \emph{cExp} is slightly lower than that of \emph{fExp} without the distillation loss constraint, which indicates the potential research value of the inherent asymmetry of the attributes. Meanwhile, without distillation, the performance of DEDN is higher than both \emph{cExp} and \emph{fExp}, demonstrating the complementary properties of the dual experts. In addition, it is worth noting that DAN removing the channel attention results in a substantial performance degradation, demonstrating the importance of channel information. Moreover, the role of $\mathcal{L}_{mal}$ in balancing the confidence of unseen and seen classes can be observed from the metrics \textbf{U} and \textbf{S}. When $\mathcal{L}_{mal}$ is removed, the metric \textbf{U} increases dramatically while \textbf{S} decreases dramatically. Finally, the results also demonstrate the importance of $\mathcal{L}_{align}$ for constraining semantic consistency.

\subsection{Empirical Analysis}

\begin{figure*}[htbp]
    \centering
    \begin{minipage}{\linewidth}
    \includegraphics[width=0.99\textwidth]{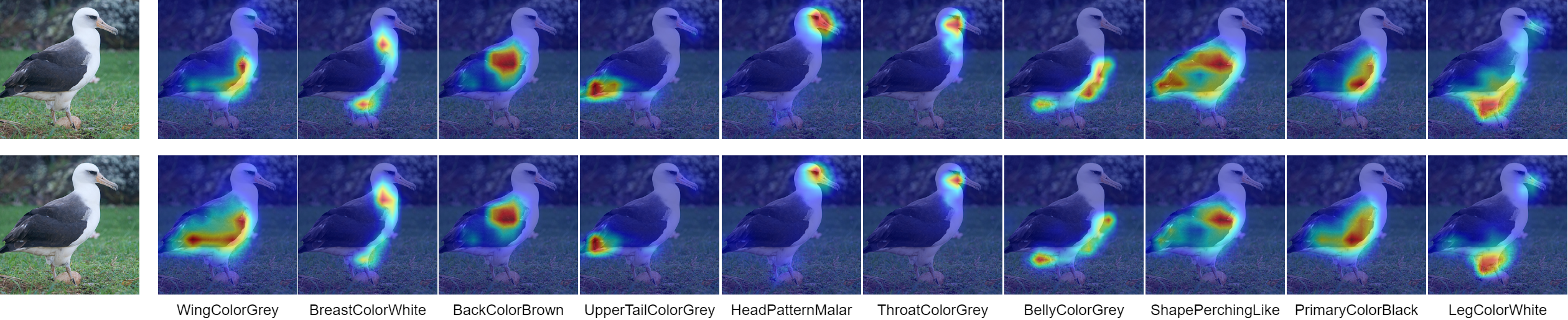}
    \end{minipage}
    \caption{Visualization of the attention heat maps. The first row represents the heat maps of \emph{cExp}, and the second row denotes the heat maps of \emph{fExp}.}
    \label{fig:attri_attention}
\end{figure*}

\begin{figure*}[htbp]
    \centering
    \begin{minipage}{0.245\linewidth}
        \centering
        \includegraphics[width=0.99\textwidth]{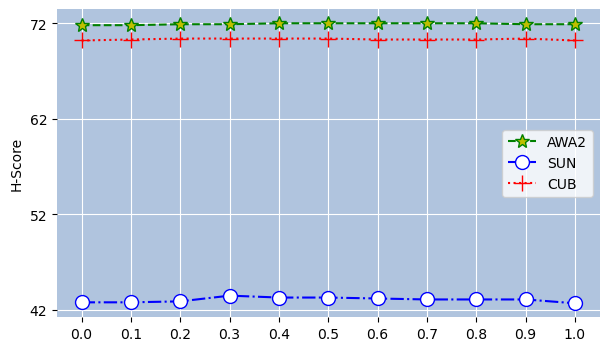}
        \centering
        (a)
    \end{minipage}
    \begin{minipage}{0.245\linewidth}
        \centering
        \includegraphics[width=0.99\textwidth]{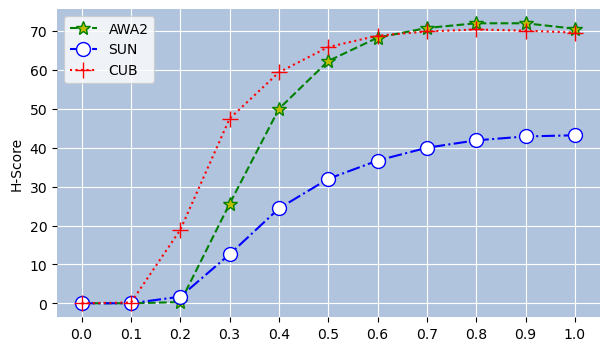}
        \centering
        (b)
    \end{minipage}
    \begin{minipage}{0.245\linewidth}
        \centering
        \includegraphics[width=0.99\textwidth]{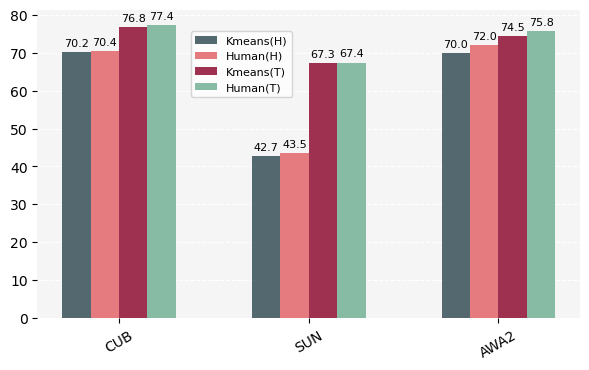}
        \centering
        (c)
    \end{minipage}
    \begin{minipage}{0.245\linewidth}
        \centering
        \includegraphics[width=0.99\textwidth]{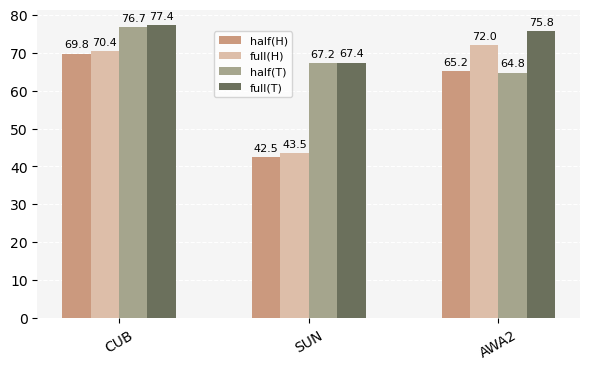}
        \centering
        (d)
    \end{minipage}
    \caption{(a) Sensitivity to $\lambda_e$. (b) Sensitivity to $\lambda_{rc}$. The harmonic mean (H) is reported. (c) Comparison with Kmeans. (d) Impact of the number of attribute clusters. The harmonic mean (H) and top-1 accuracy (T) are reported.}
    \label{fig:empirical_analysis}
\end{figure*}

\subsection{The influence of parameters $\lambda_{e}$ and $\lambda_{rc}$}
We launch a series of empirical analyses, including evaluating the impact of parameters $\lambda_{e}$ and $\lambda_{rc}$ on the final performance. Figure \ref{fig:empirical_analysis} (a) illustrates the sensitivity of the harmonic mean for each dataset with respect to parameter $\lambda_{e}$. It can be observed that the influence of parameter a is extremely small. Of particular note, when $\lambda_{e}$ is set to $1$ or $0$, it indicates that only the \emph{cExp} or \emph{fExp} after distillation learning is used for the inference phase. It implies that by mutual distillation learning, each of the two experts learns the strengths of the other, thereby reaching an agreement. Figure \ref{fig:empirical_analysis} (b) illustrates the impact of $\lambda_{rc}$. It can be seen that setting $\lambda_{rc}$ above $0.7$ stabilizes the performance. Optimization is achieved when it is set between $0.7$ and $0.9$.

\subsubsection{The influence of different clustering algorithms}
We further evaluate the impact of the clustering algorithm on performance. In Introducing Table \ref{tab:attribute_cluster}, we have explained that attribute clusters are obtained by humans to classify the attribute sets based on their characteristics. In this subsection, we use the K-Means algorithm for attribute clustering as a comparison to evaluate the performance. The experimental results are shown in Figure \ref{fig:empirical_analysis} (c), where the harmonic mean (H) and top-1 accuracy (T) are reported. From the figure, it can be seen that the K-Means algorithm is slightly poorer compared to human classification, but a good result is also achieved. It again shows that the idea of dividing the attribute set into different clusters holds great promise.

\subsubsection{The influence of the number of attribute clusters}
We evaluate the impact of the number of attribute clusters on performance. The attributes of CUB, SUN, and AWA2 are classified into $6$, $4$, and $4$ categories, respectively (Table \ref{tab:attribute_cluster}). In this subsection, we halve the categories, i.e., the numbers of attribute clusters for CUB, SUN, and AWA2 are $3$, $2$, and $2$. The experimental results are shown in Figure \ref{fig:empirical_analysis} (d), where \emph{half} denotes that the cluster number is halved. We can see that \emph{half} leads to a reduction of \textbf{H} by 0.6\%, 1.0\%, and 6.8\%, respectively, and a reduction of \textbf{T} by 0.7\%, 0.2\%, and 11\%, respectively. The results show that detailed attribute classification facilitates the model in capturing more fine-grained information and thus improves the performance.

\subsubsection{Visual analysis of attention.}
We perform a visual analysis of the attention of the two experts, and the schematic is shown in Figure \ref{fig:attri_attention}. It can be observed that \emph{cExp} has a better localization for some global attributes, such as \emph{HeadPatternMaler, BellyColorGrey, ShapePerchingLike}. Meanwhile, \emph{fExp} has more detailed and precise localization for some local attributes, such as \emph{UpperTailColorGrey, ThroatColorGrey, LegColorWhite}. The two experts collaborate and learn in a complementary way to improve together, which leads to better performance.

\section{Conclusion}

In this paper, we analyze the impact of attribute annotations and channel information on the regional visual-attribute modeling task. We argue that the intrinsic asymmetry of attributes is one of the important bottlenecks constraining existing approaches and propose a simple yet effective framework named DEDN to address this problem. DEDN consists of two expert networks, one with complete attribute-domain perception to harmonize the global correlation confidence and the other consisting of multiple subnetworks, each focusing on a specific attribute domain to capture fine-grained association information. Both of them complement each other and learn cooperatively. Meanwhile, we introduce DAN as a strong backbone, a novel attention network that incorporates both region and channel knowledge. Moreover, we present a new loss named MAL to train the network. Numerous experiments demonstrate the significant superiority of the proposed approach.

\bibliographystyle{named}
\bibliography{ijcai24}

\begin{thebibliography}{}

\bibitem[\protect\citeauthoryear{Annadani and Biswas}{2018}]{annadani2018preserving}
Yashas Annadani and Soma Biswas.
\newblock Preserving semantic relations for zero-shot learning.
\newblock In {\em Proceedings of the IEEE Conference on Computer Vision and Pattern Recognition}, pages 7603--7612, 2018.

\bibitem[\protect\citeauthoryear{Changpinyo \bgroup \em et al.\egroup }{2016}]{changpinyo2016synthesized}
Soravit Changpinyo, Wei-Lun Chao, Boqing Gong, and Fei Sha.
\newblock Synthesized classifiers for zero-shot learning.
\newblock In {\em Proceedings of the IEEE Conference on Computer Vision and Pattern Recognition}, pages 5327--5336, 2016.

\bibitem[\protect\citeauthoryear{Chao \bgroup \em et al.\egroup }{2016}]{chao2016empirical}
Wei-Lun Chao, Soravit Changpinyo, Boqing Gong, and Fei Sha.
\newblock An empirical study and analysis of generalized zero-shot learning for object recognition in the wild.
\newblock In {\em Computer Vision--ECCV 2016: 14th European Conference, Amsterdam, The Netherlands, October 11-14, 2016, Proceedings, Part II 14}, pages 52--68. Springer, 2016.

\bibitem[\protect\citeauthoryear{Chen \bgroup \em et al.\egroup }{2021a}]{chen2021free}
Shiming Chen, Wenjie Wang, Beihao Xia, Qinmu Peng, Xinge You, Feng Zheng, and Ling Shao.
\newblock Free: Feature refinement for generalized zero-shot learning.
\newblock In {\em Proceedings of the IEEE/CVF International Conference on Computer Vision}, pages 122--131, 2021.

\bibitem[\protect\citeauthoryear{Chen \bgroup \em et al.\egroup }{2021b}]{chen2021hsva}
Shiming Chen, Guosen Xie, Yang Liu, Qinmu Peng, Baigui Sun, Hao Li, Xinge You, and Ling Shao.
\newblock Hsva: Hierarchical semantic-visual adaptation for zero-shot learning.
\newblock {\em Advances in Neural Information Processing Systems}, 34:16622--16634, 2021.

\bibitem[\protect\citeauthoryear{Chen \bgroup \em et al.\egroup }{2021c}]{chen2021semantics}
Zhi Chen, Yadan Luo, Ruihong Qiu, Sen Wang, Zi~Huang, Jingjing Li, and Zheng Zhang.
\newblock Semantics disentangling for generalized zero-shot learning.
\newblock In {\em Proceedings of the IEEE/CVF International Conference on Computer Vision}, pages 8712--8720, 2021.

\bibitem[\protect\citeauthoryear{Chen \bgroup \em et al.\egroup }{2022a}]{chen2022transzero}
Shiming Chen, Ziming Hong, Yang Liu, Guo-Sen Xie, Baigui Sun, Hao Li, Qinmu Peng, Ke~Lu, and Xinge You.
\newblock Transzero: Attribute-guided transformer for zero-shot learning.
\newblock In {\em Proceedings of the AAAI Conference on Artificial Intelligence}, volume~36, pages 330--338, 2022.

\bibitem[\protect\citeauthoryear{Chen \bgroup \em et al.\egroup }{2022b}]{chen2022msdn}
Shiming Chen, Ziming Hong, Guo-Sen Xie, Wenhan Yang, Qinmu Peng, Kai Wang, Jian Zhao, and Xinge You.
\newblock Msdn: Mutually semantic distillation network for zero-shot learning.
\newblock In {\em Proceedings of the IEEE/CVF Conference on Computer Vision and Pattern Recognition}, pages 7612--7621, 2022.

\bibitem[\protect\citeauthoryear{Felix \bgroup \em et al.\egroup }{2018}]{felix2018multi}
Rafael Felix, Ian Reid, Gustavo Carneiro, et~al.
\newblock Multi-modal cycle-consistent generalized zero-shot learning.
\newblock In {\em Proceedings of the European Conference on Computer Vision}, pages 21--37, 2018.

\bibitem[\protect\citeauthoryear{Guo \bgroup \em et al.\egroup }{2023}]{guo2023graph}
Jingcai Guo, Song Guo, Qihua Zhou, Ziming Liu, Xiaocheng Lu, and Fushuo Huo.
\newblock Graph knows unknowns: Reformulate zero-shot learning as sample-level graph recognition.
\newblock In {\em Proceedings of the AAAI Conference on Artificial Intelligence}, volume~37, pages 7775--7783, 2023.

\bibitem[\protect\citeauthoryear{Han \bgroup \em et al.\egroup }{2021}]{han2021contrastive}
Zongyan Han, Zhenyong Fu, Shuo Chen, and Jian Yang.
\newblock Contrastive embedding for generalized zero-shot learning.
\newblock In {\em Proceedings of the IEEE/CVF Conference on Computer Vision and Pattern Recognition}, pages 2371--2381, 2021.

\bibitem[\protect\citeauthoryear{He \bgroup \em et al.\egroup }{2016}]{he2016deep}
Kaiming He, Xiangyu Zhang, Shaoqing Ren, and Jian Sun.
\newblock Deep residual learning for image recognition.
\newblock In {\em Proceedings of the IEEE Conference on Computer Vision and Pattern Recognition}, pages 770--778, 2016.

\bibitem[\protect\citeauthoryear{Huynh and Elhamifar}{2020}]{huynh2020fine}
Dat Huynh and Ehsan Elhamifar.
\newblock Fine-grained generalized zero-shot learning via dense attribute-based attention.
\newblock In {\em Proceedings of the IEEE/CVF Conference on Computer Vision and Pattern Recognition}, pages 4483--4493, 2020.

\bibitem[\protect\citeauthoryear{Jiang \bgroup \em et al.\egroup }{2019}]{jiang2019transferable}
Huajie Jiang, Ruiping Wang, Shiguang Shan, and Xilin Chen.
\newblock Transferable contrastive network for generalized zero-shot learning.
\newblock In {\em Proceedings of the IEEE/CVF International Conference on Computer Vision}, pages 9765--9774, 2019.

\bibitem[\protect\citeauthoryear{Keshari \bgroup \em et al.\egroup }{2020}]{keshari2020generalized}
Rohit Keshari, Richa Singh, and Mayank Vatsa.
\newblock Generalized zero-shot learning via over-complete distribution.
\newblock In {\em Proceedings of the IEEE/CVF Conference on Computer Vision and Pattern Recognition}, pages 13300--13308, 2020.

\bibitem[\protect\citeauthoryear{Kodirov \bgroup \em et al.\egroup }{2017}]{kodirov2017semantic}
Elyor Kodirov, Tao Xiang, and Shaogang Gong.
\newblock Semantic autoencoder for zero-shot learning.
\newblock In {\em Proceedings of the IEEE Conference on Computer Vision and Pattern Recognition}, pages 3174--3183, 2017.

\bibitem[\protect\citeauthoryear{Kong \bgroup \em et al.\egroup }{2022}]{kong2022compactness}
Xia Kong, Zuodong Gao, Xiaofan Li, Ming Hong, Jun Liu, Chengjie Wang, Yuan Xie, and Yanyun Qu.
\newblock En-compactness: Self-distillation embedding \& contrastive generation for generalized zero-shot learning.
\newblock In {\em Proceedings of the IEEE/CVF Conference on Computer Vision and Pattern Recognition}, pages 9306--9315, 2022.

\bibitem[\protect\citeauthoryear{Lampert \bgroup \em et al.\egroup }{2009}]{lampert2009learning}
Christoph~H Lampert, Hannes Nickisch, and Stefan Harmeling.
\newblock Learning to detect unseen object classes by between-class attribute transfer.
\newblock In {\em 2009 IEEE Conference on Computer Vision and Pattern Recognition}, pages 951--958. IEEE, 2009.

\bibitem[\protect\citeauthoryear{Li \bgroup \em et al.\egroup }{2019}]{li2019leveraging}
Jingjing Li, Mengmeng Jing, Ke~Lu, Zhengming Ding, Lei Zhu, and Zi~Huang.
\newblock Leveraging the invariant side of generative zero-shot learning.
\newblock In {\em Proceedings of the IEEE/CVF Conference on Computer Vision and Pattern Recognition}, pages 7402--7411, 2019.

\bibitem[\protect\citeauthoryear{Li \bgroup \em et al.\egroup }{2021}]{li2021generalized}
Xiangyu Li, Zhe Xu, Kun Wei, and Cheng Deng.
\newblock Generalized zero-shot learning via disentangled representation.
\newblock In {\em Proceedings of the AAAI Conference on Artificial Intelligence}, volume~35, pages 1966--1974, 2021.

\bibitem[\protect\citeauthoryear{Li \bgroup \em et al.\egroup }{2023}]{Li2023boosting}
Xiaofan Li, Yachao Zhang, Shiran Bian, Yanyun Qu, Yuan Xie, Zhongchao Shi, and Jianping Fan.
\newblock Vs-boost: Boosting visual-semantic association for generalized zero-shot learning.
\newblock In {\em International Joint Conference on Artificial Intelligence}, 2023.

\bibitem[\protect\citeauthoryear{Liu \bgroup \em et al.\egroup }{2018}]{liu2018generalized}
Shichen Liu, Mingsheng Long, Jianmin Wang, and Michael~I Jordan.
\newblock Generalized zero-shot learning with deep calibration network.
\newblock {\em Advances in Neural Information Processing Systems}, 31, 2018.

\bibitem[\protect\citeauthoryear{Liu \bgroup \em et al.\egroup }{2019}]{liu2019attribute}
Yang Liu, Jishun Guo, Deng Cai, and Xiaofei He.
\newblock Attribute attention for semantic disambiguation in zero-shot learning.
\newblock In {\em Proceedings of the IEEE/CVF International Conference on Computer Vision}, pages 6698--6707, 2019.

\bibitem[\protect\citeauthoryear{Liu \bgroup \em et al.\egroup }{2020}]{liu2020attribute}
Lu~Liu, Tianyi Zhou, Guodong Long, Jing Jiang, and Chengqi Zhang.
\newblock Attribute propagation network for graph zero-shot learning.
\newblock In {\em Proceedings of the AAAI Conference on Artificial Intelligence}, volume~34, pages 4868--4875, 2020.

\bibitem[\protect\citeauthoryear{Narayan \bgroup \em et al.\egroup }{2020}]{narayan2020latent}
Sanath Narayan, Akshita Gupta, Fahad~Shahbaz Khan, Cees~GM Snoek, and Ling Shao.
\newblock Latent embedding feedback and discriminative features for zero-shot classification.
\newblock In {\em Computer Vision--ECCV 2020: 16th European Conference, Glasgow, UK, August 23--28, 2020, Proceedings, Part XXII 16}, pages 479--495. Springer, 2020.

\bibitem[\protect\citeauthoryear{Patterson and Hays}{2012}]{patterson2012sun}
Genevieve Patterson and James Hays.
\newblock Sun attribute database: Discovering, annotating, and recognizing scene attributes.
\newblock In {\em 2012 IEEE Conference on Computer Vision and Pattern Recognition}, pages 2751--2758. IEEE, 2012.

\bibitem[\protect\citeauthoryear{Radford \bgroup \em et al.\egroup }{2018}]{radford2018improving}
Alec Radford, Karthik Narasimhan, Tim Salimans, Ilya Sutskever, et~al.
\newblock Improving language understanding by generative pre-training.
\newblock 2018.

\bibitem[\protect\citeauthoryear{Schonfeld \bgroup \em et al.\egroup }{2019}]{schonfeld2019generalized}
Edgar Schonfeld, Sayna Ebrahimi, Samarth Sinha, Trevor Darrell, and Zeynep Akata.
\newblock Generalized zero-and few-shot learning via aligned variational autoencoders.
\newblock In {\em Proceedings of the IEEE/CVF Conference on Computer Vision and Pattern Recognition}, pages 8247--8255, 2019.

\bibitem[\protect\citeauthoryear{Verma \bgroup \em et al.\egroup }{2018}]{verma2018generalized}
Vinay~Kumar Verma, Gundeep Arora, Ashish Mishra, and Piyush Rai.
\newblock Generalized zero-shot learning via synthesized examples.
\newblock In {\em Proceedings of the IEEE Conference on Computer Vision and Pattern Recognition}, pages 4281--4289, 2018.

\bibitem[\protect\citeauthoryear{Vyas \bgroup \em et al.\egroup }{2020}]{vyas2020leveraging}
Maunil~R Vyas, Hemanth Venkateswara, and Sethuraman Panchanathan.
\newblock Leveraging seen and unseen semantic relationships for generative zero-shot learning.
\newblock In {\em Computer Vision--ECCV 2020: 16th European Conference, Glasgow, UK, August 23--28, 2020, Proceedings, Part XXX 16}, pages 70--86. Springer, 2020.

\bibitem[\protect\citeauthoryear{Wah \bgroup \em et al.\egroup }{2011}]{wah2011caltech}
Catherine Wah, Steve Branson, Peter Welinder, Pietro Perona, and Serge Belongie.
\newblock The caltech-ucsd birds-200-2011 dataset.
\newblock 2011.

\bibitem[\protect\citeauthoryear{Wang \bgroup \em et al.\egroup }{2021}]{wang2021dual}
Chaoqun Wang, Shaobo Min, Xuejin Chen, Xiaoyan Sun, and Houqiang Li.
\newblock Dual progressive prototype network for generalized zero-shot learning.
\newblock {\em Advances in Neural Information Processing Systems}, 34:2936--2948, 2021.

\bibitem[\protect\citeauthoryear{Xian \bgroup \em et al.\egroup }{2017}]{xian2017zero}
Yongqin Xian, Bernt Schiele, and Zeynep Akata.
\newblock Zero-shot learning-the good, the bad and the ugly.
\newblock In {\em Proceedings of the IEEE Conference on Computer Vision and Pattern Recognition}, pages 4582--4591, 2017.

\bibitem[\protect\citeauthoryear{Xian \bgroup \em et al.\egroup }{2018}]{xian2018feature}
Yongqin Xian, Tobias Lorenz, Bernt Schiele, and Zeynep Akata.
\newblock Feature generating networks for zero-shot learning.
\newblock In {\em Proceedings of the IEEE Conference on Computer Vision and Pattern Recognition}, pages 5542--5551, 2018.

\bibitem[\protect\citeauthoryear{Xian \bgroup \em et al.\egroup }{2019}]{xian2019f}
Yongqin Xian, Saurabh Sharma, Bernt Schiele, and Zeynep Akata.
\newblock f-vaegan-d2: A feature generating framework for any-shot learning.
\newblock In {\em Proceedings of the IEEE/CVF Conference on Computer Vision and Pattern Recognition}, pages 10275--10284, 2019.

\bibitem[\protect\citeauthoryear{Xie \bgroup \em et al.\egroup }{2019}]{xie2019attentive}
Guo-Sen Xie, Li~Liu, Xiaobo Jin, Fan Zhu, Zheng Zhang, Jie Qin, Yazhou Yao, and Ling Shao.
\newblock Attentive region embedding network for zero-shot learning.
\newblock In {\em Proceedings of the IEEE/CVF Conference on Computer Vision and Pattern Recognition}, pages 9384--9393, 2019.

\bibitem[\protect\citeauthoryear{Xie \bgroup \em et al.\egroup }{2020}]{xie2020region}
Guo-Sen Xie, Li~Liu, Fan Zhu, Fang Zhao, Zheng Zhang, Yazhou Yao, Jie Qin, and Ling Shao.
\newblock Region graph embedding network for zero-shot learning.
\newblock In {\em Computer Vision--ECCV 2020: 16th European Conference, Glasgow, UK, August 23--28, 2020, Proceedings, Part IV 16}, pages 562--580. Springer, 2020.

\bibitem[\protect\citeauthoryear{Xie \bgroup \em et al.\egroup }{2022}]{xie2022leveraging}
Guo-Sen Xie, Xu-Yao Zhang, Tian-Zhu Xiang, Fang Zhao, Zheng Zhang, Ling Shao, and Xuelong Li.
\newblock Leveraging balanced semantic embedding for generative zero-shot learning.
\newblock {\em IEEE Transactions on Neural Networks and Learning Systems}, 2022.

\bibitem[\protect\citeauthoryear{Xu \bgroup \em et al.\egroup }{2020}]{xu2020attribute}
Wenjia Xu, Yongqin Xian, Jiuniu Wang, Bernt Schiele, and Zeynep Akata.
\newblock Attribute prototype network for zero-shot learning.
\newblock {\em Advances in Neural Information Processing Systems}, 33:21969--21980, 2020.

\bibitem[\protect\citeauthoryear{Yu \bgroup \em et al.\egroup }{2020}]{yu2020episode}
Yunlong Yu, Zhong Ji, Jungong Han, and Zhongfei Zhang.
\newblock Episode-based prototype generating network for zero-shot learning.
\newblock In {\em Proceedings of the IEEE/CVF Conference on Computer Vision and Pattern Recognition}, pages 14035--14044, 2020.

\bibitem[\protect\citeauthoryear{Zhang \bgroup \em et al.\egroup }{2017}]{zhang2017learning}
Li~Zhang, Tao Xiang, and Shaogang Gong.
\newblock Learning a deep embedding model for zero-shot learning.
\newblock In {\em Proceedings of the IEEE Conference on Computer Vision and Pattern Recognition}, pages 2021--2030, 2017.

\bibitem[\protect\citeauthoryear{Zhu \bgroup \em et al.\egroup }{2019}]{zhu2019semantic}
Yizhe Zhu, Jianwen Xie, Zhiqiang Tang, Xi~Peng, and Ahmed Elgammal.
\newblock Semantic-guided multi-attention localization for zero-shot learning.
\newblock {\em Advances in Neural Information Processing Systems}, 32, 2019.

\end{thebibliography}

\end{document}